\documentclass[runningheads]{llncs}
\usepackage[T1]{fontenc}
\usepackage{graphicx}
\usepackage{amsmath, amssymb}
\usepackage{algorithm}
\usepackage{algorithmic}
\usepackage{booktabs}
\usepackage{tabularx}
\usepackage{rotating}
\begin{document}

\title{Medication-Aware Financial Exploitation Detection for Alzheimer's Patients Using Edge-Aware Interaction Risk Modeling}
\titlerunning{Medication-Aware Exploitation Detection}

\author{
Farzana Akter\inst{1} \and
Lisan Al Amin\inst{1} \and
Rakib Hossain\inst{1} \and
Chaitanya Gunupudi\inst{2} \and
Faisal Quader\inst{2}\thanks{Corresponding author: \email{quader@umd.edu}}
}

\authorrunning{F. Akter et al.}

\institute{
Cognitive Links LLC, Maryland, USA\\
\email{\{farzana.akter9814,lisanalamin,rakib.sat18\}@gmail.com}
\and
University of Maryland, College Park, USA\\
\email{chaitanya.gunupudi970@gmail.com, quader@umd.edu}
}

\maketitle
\begin{abstract}
Financial exploitation is a growing concern for people with Alzheimer’s disease, especially during periods of reduced cognitive stability. Conventional fraud detection systems usually rely on financial behavior alone and ignore clinically relevant factors that may alter vulnerability. This paper proposes a medication-aware framework that synchronizes medication adherence with transaction-level monitoring to improve detection of cognitively risky financial events. A hybrid simulation dataset was constructed for 180 patients across 45 days, producing 8{,}100 medication records and 30{,}855 transactions. The framework evaluates amount anomaly, vendor novelty, transaction frequency, time deviation, and medication adherence through financial-only, additive medication-aware, and interaction-aware logistic models. Results show that the financial-only baseline obtained the highest global F1-score of 0.5000, but the interaction-aware model improved recall during medication-induced vulnerability windows from 0.7442 to 0.9070 and achieved the highest average precision for ranked high-risk cases. The findings suggest that medication adherence is most useful as a contextual modifier of financial risk rather than as an isolated predictor.

\keywords{Alzheimer's disease \and financial exploitation detection \and medication adherence \and vulnerability-aware risk modeling \and edge-aware healthcare analytics}
\end{abstract}

\section{Introduction}

Financial exploitation is a major threat for older adults in digital banking and remote transaction environments. Recent work shows that financial harm in later life is shaped by both external deception and age-related changes in cognition, affect, and judgment \cite{_1}. This concern is amplified in Alzheimer’s disease, where impaired executive function and declining self-monitoring can reduce a patient’s ability to recognize abnormal transactions or respond to manipulation.

Cognitive decline can also affect financial capacity before patients fully recognize the decline themselves. Mazzonna and Peracchi showed that severe cognitive decline combined with poor awareness is associated with meaningful financial losses \cite{_2}. This implies that exploitation detection for Alzheimer’s patients should not be treated only as a standard anomaly detection problem. It should also consider temporally changing vulnerability, because the same transaction may deserve different interpretation when the patient is clinically unstable.

Digital dementia-care research has made progress in voice assistance, personalization, smart assistants, and remote monitoring \cite{_3,_4,_5,_6}. Privacy and governance studies further show that dementia-related monitoring must minimize unnecessary exposure of sensitive personal information \cite{_7,_12}. However, these advances do not directly solve the financial protection problem. Existing systems often focus on communication, caregiving, behavioral monitoring, or general agentic support rather than medication-aware transaction risk modeling \cite{_8,_9,_10}. At the same time, research on dementia and financial capacity confirms the importance of early financial markers but rarely integrates dynamic clinical context into transaction-level detection \cite{_13,_14,_16}.

This paper addresses this gap by developing a simulation-based, medication-aware framework for financial exploitation detection. Medication non-adherence is modeled as a dynamic vulnerability signal, not as a static patient descriptor. The study asks whether financial anomalies become more informative when interpreted during missed-medication windows. The objective is not to claim live clinical or banking deployment, but to test whether medication-aware interaction modeling improves vulnerability-sensitive detection compared with financial-only and additive baselines.

The contributions are fourfold. First, the paper introduces a medication-aware risk modeling framework that treats adherence as a contextual modifier of financial risk. Second, it constructs a synchronized hybrid dataset linking daily medication events with transaction streams. Third, it compares interpretable weighted scoring with logistic baselines and an interaction-aware model. Fourth, it evaluates performance both globally and inside medication-induced vulnerability windows, where missed exploitation events are most clinically concerning.

\section{Related Work and Research Gap}

Financial exploitation among older adults has been studied across fraud, aging, cognition, and decision making. Ebner et al. showed that older adults may face deception risks because of changes in social judgment and affective processing \cite{_1}. Studies of cognitive decline and financial capacity further suggest that financial behavior can reveal early functional deterioration and neurocognitive risk \cite{_2,_13,_14}. Uncertainty-aware financial risk modeling also supports the need for defensible and interpretable risk estimates in compliance-sensitive settings \cite{al2025bayesian}. Recent reviews confirm that financial exploitation among people living with dementia is a growing problem requiring preventive intervention \cite{_16}. However, most existing work does not treat medication adherence as a time-varying vulnerability signal.

Prior studies often model vulnerability as a stable person-level condition, although Alzheimer’s-related risk may fluctuate with medication routines, fatigue, sleep, caregiver availability, and daily disruption. Medication adherence is therefore clinically relevant because missed medication can indicate reduced routine stability or periods of cognitive vulnerability. Olchanski et al. showed that Alzheimer’s medication use and adherence vary across patient populations \cite{_11}. Related machine learning work on opioid patient classification also demonstrates that medication-linked clinical data can support meaningful risk stratification \cite{al2022data}. In this context, a missed dose may not directly cause fraud, but it can change how moderate financial anomalies should be interpreted.

AI-supported dementia-care systems provide useful technical context. LLM-based and memory-coordinated assistants have been proposed for communication and personalization in older-adult care \cite{_3,_4}. Co-designed smart assistants, context-aware mobile systems, and remote monitoring reviews also show that AI-enabled dementia support is increasingly feasible \cite{_5,_6,_8}. Risk-controlled AI support for vulnerable users reinforces the need for safety-aware design, interpretability, and human oversight \cite{hossain2025risk}. Agentic dementia-care frameworks suggest broader coordination possibilities, but they remain architectural rather than empirical studies of medication-aware financial risk \cite{_9,_10}.

Privacy-aware monitoring is also essential. Privacy-preserving clinical learning, including hybrid federated and split learning, provides a relevant path for sensitive patient-centered prediction without unrestricted centralized data exposure \cite{akter2026hybrid}. Reviews of IoMT privacy in Alzheimer’s care emphasize consent, governance, and data minimization \cite{_7}, while privacy-protecting video-based risk detection shows that useful monitoring should not require full exposure of raw personal data \cite{_12}. Participatory work on deepfake scams and older adults further shows that protection should preserve autonomy and dignity rather than simply block activity \cite{_15}. These principles support an edge-aware design in which medication and transaction signals are converted into limited risk features before escalation.

The research gap is therefore clear: prior work links dementia to financial vulnerability and supports context-aware dementia monitoring, but it rarely tests whether medication adherence improves transaction-level exploitation detection. This study addresses that gap by comparing financial-only, additive medication-aware, and interaction-aware formulations under controlled simulation.
\section{Methodology}

\subsection{Framework Overview}

\begin{figure}[htbp]
    \centering
    \includegraphics[width=\linewidth]{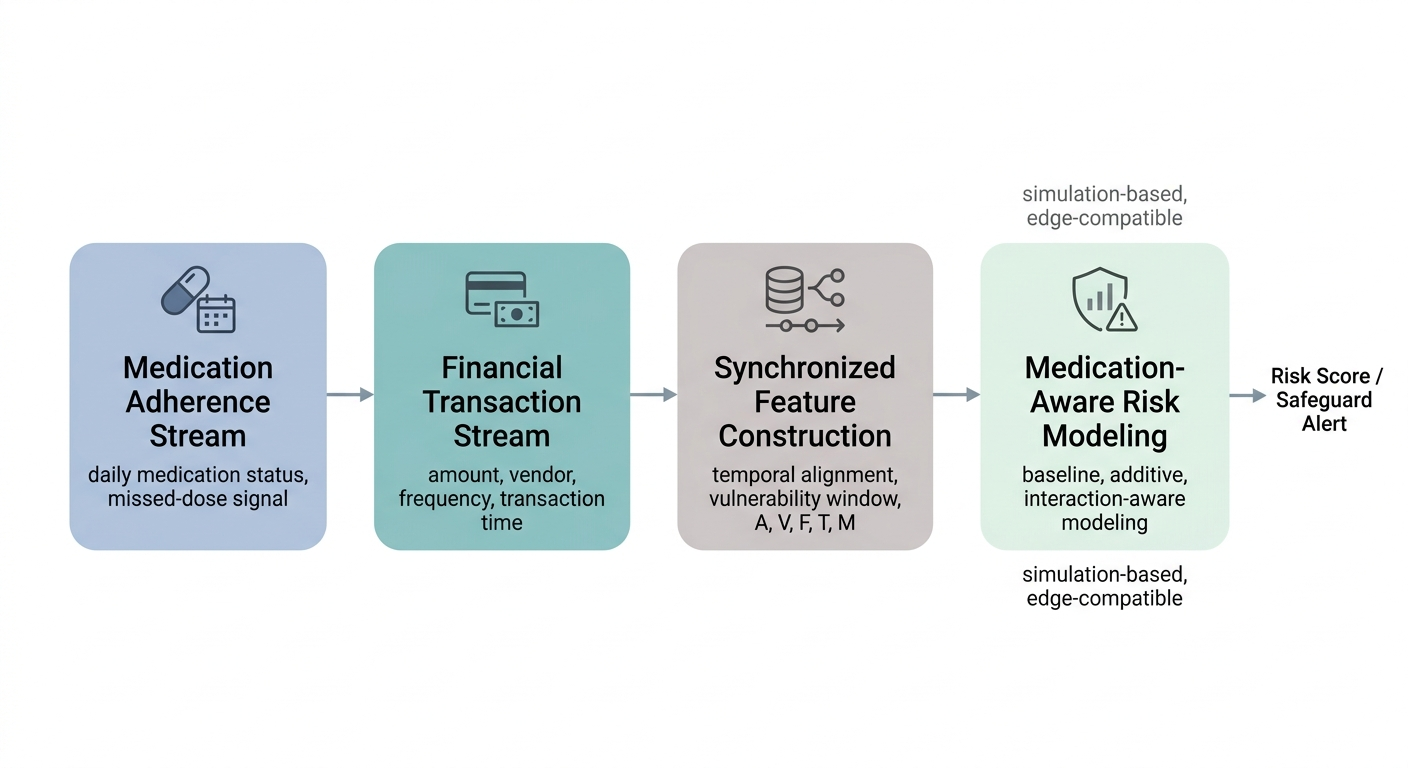}
    \caption{Proposed medication-aware financial exploitation detection framework. Medication adherence and transaction streams are temporally synchronized to construct clinically informed financial risk representations. The framework compares baseline, additive medication-aware, and interaction-aware models, then produces vulnerability-sensitive risk scores and prioritized safeguard actions in a simulation-based, edge-compatible setting.}
    \label{fig:framework}
\end{figure}

The proposed framework synchronizes medication adherence records with financial transactions for Alzheimer’s patients. It contains three layers: a perception layer that captures medication and transaction events, a reasoning layer that converts them into aligned risk features, and an action layer that produces risk scores for review or escalation. The framework is edge-aware because it uses compact features rather than raw clinical or banking histories, which is consistent with privacy-sensitive dementia monitoring principles \cite{_7,_12}. Figure~\ref{fig:framework} shows the overall architecture.

\subsection{Hybrid Dataset Construction}

A synchronized hybrid dataset was constructed with 180 synthetic patients observed over 45 days, producing 8{,}100 medication records and 30{,}855 financial transactions. Each patient profile included age, cognitive-risk level, adherence probability, expected transaction rate, typical spending level, and familiar vendors. Medication was modeled as a daily scheduled event. A taken dose indicated stable state, while a missed dose created a vulnerability window beginning two hours after the scheduled time and lasting twelve hours. This reflects the hypothesis that suspicious financial behavior should be interpreted more cautiously during clinically elevated vulnerability \cite{_11,_13}.

Transactions were generated from patient-specific Poisson processes with regular and burst-like activity. Burst activity was made more likely on missed-medication days. For each transaction, the simulator assigned amount, vendor identity, vendor novelty, recent one-hour transaction count, timing risk, vulnerability-window status, and class label. Labeling was probabilistic rather than deterministic: exploitation probability increased during vulnerability windows and also depended on unfamiliar vendors, high amounts, local bursts, and high-risk timing. Non-cognitive anomalies were generated independently of medication status so that suspicious financial activity could occur outside vulnerability windows. Algorithm~\ref{alg:dataset_construction} summarizes the construction process.

\begin{algorithm}[htbp]
\caption{Construction of the Synchronized Hybrid Dataset}
\label{alg:dataset_construction}
\begin{algorithmic}[1]
\STATE Initialize patient cohort $\mathcal{P}$
\FOR{each patient $p \in \mathcal{P}$}
    \STATE Sample patient-specific behavioral profile
    \FOR{each day $d$ in the observation horizon}
        \STATE Generate scheduled medication event
        \IF{dose is missed}
            \STATE Define vulnerability window $[t_d^{start}, t_d^{end}]$
        \ENDIF
        \STATE Sample daily transaction count
        \STATE Generate timestamped transactions with possible burst behavior
        \FOR{each transaction $x_i$}
            \STATE Compute financial and timing features
            \STATE Check vulnerability-window membership
            \STATE Assign probabilistic label from clinical-financial context
            \STATE Store synchronized record
        \ENDFOR
    \ENDFOR
\ENDFOR
\STATE Return dataset $\mathcal{D}$
\end{algorithmic}
\end{algorithm}

\subsection{Feature Engineering and Risk Formulation}

For each transaction $x_i$, five features were extracted: amount anomaly $A_i$, vendor novelty $V_i$, frequency spike $F_i$, time deviation $T_i$, and medication adherence $M_i$. Medication adherence was encoded as
\begin{equation}
M_i =
\begin{cases}
1, & \text{if the scheduled medication dose was taken}, \\ 
0, & \text{if the scheduled medication dose was missed}.
\end{cases}
\end{equation}
The missed-medication indicator used for interaction modeling was $1-M_i$.

The amount anomaly feature used a robust positive deviation score based on the patient-specific median and median absolute deviation:
\begin{equation}
z_i^{(A)}=\frac{0.6745(a_i-\mathrm{median}(A_p))}{\mathrm{MAD}(A_p)+\epsilon}, \quad
A_i=1-\exp[-\max(0,z_i^{(A)})].
\end{equation}
Vendor novelty was binary:
\begin{equation}
V_i=
\begin{cases}
1, & \text{if vendor}(x_i)\notin \mathcal{V}_p,\\
0, & \text{otherwise},
\end{cases}
\end{equation}
where $\mathcal{V}_p$ is the familiar vendor set for patient $p$. Frequency spike used the one-hour transaction count $c_i$:
\begin{equation}
z_i^{(F)}=\frac{c_i-\mu_p^{(F)}}{\sigma_p^{(F)}+\epsilon}, \quad
F_i=1-\exp[-\max(0,z_i^{(F)})].
\end{equation}
Time deviation was encoded by hour of day:
\begin{equation}
T_i=
\begin{cases}
1.0, & 0\leq h_i\leq5,\\
0.6, & 6\leq h_i\leq7\text{ or }20\leq h_i\leq23,\\
0.1, & \text{otherwise}.
\end{cases}
\end{equation}

An interpretable medication-aware score was defined as
\begin{equation}
R_i=(1-M_i)(0.40A_i+0.25V_i+0.20F_i+0.15T_i),
\label{eq:risk_score}
\end{equation}
where the score becomes active only when medication is missed. A transaction was flagged when $R_i\geq \tau_R$. This rule-based model served as an interpretable reference rather than the primary learned detector.

\subsection{Predictive Modeling and Evaluation}

Three logistic formulations were evaluated. The financial-only baseline used $(A_i,V_i,F_i,T_i)$. The additive medication-aware model added $M_i$. The interaction-aware model retained the original features and added $A_i(1-M_i)$, $V_i(1-M_i)$, $F_i(1-M_i)$, and $T_i(1-M_i)$:
\begin{equation}
\begin{aligned}
P(y_i=1|\mathbf{x}_i)=\sigma(&\beta_0+\beta_AA_i+\beta_VV_i+\beta_FF_i+\beta_TT_i+\beta_MM_i \\
&+\beta_{AM}A_i(1-M_i)+\beta_{VM}V_i(1-M_i) \\
&+\beta_{FM}F_i(1-M_i)+\beta_{TM}T_i(1-M_i)).
\end{aligned}
\end{equation}
This formulation directly tests whether missed medication changes the interpretation of financial anomalies. The evaluation framing is also consistent with calibrated and temporally validated prediction under operational constraints, where model usefulness depends on reliable deployment behavior rather than accuracy alone \cite{hossain2025profit}. The end-to-end detection logic is summarized in Algorithm~\ref{alg:detection}.

\begin{algorithm}[htbp]
\caption{Medication-Aware Cognitive Exploitation Detection}
\label{alg:detection}
\begin{algorithmic}[1]
\REQUIRE Transaction $x_i$ with features $(A_i,V_i,F_i,T_i,M_i)$
\STATE Compute medication-aware risk score $R_i$ using Eq.~\ref{eq:risk_score}
\STATE Construct interaction terms $A_i(1-M_i),V_i(1-M_i),F_i(1-M_i),T_i(1-M_i)$
\STATE Estimate exploitation probability $\hat{p}_i$ with the selected model
\IF{$\hat{p}_i\geq\tau^*$}
    \STATE Flag transaction as cognitive exploitation risk
\ELSE
    \STATE Mark transaction as non-exploitative
\ENDIF
\RETURN predicted label $\hat{y}_i$
\end{algorithmic}
\end{algorithm}

To avoid patient identity leakage, the data were split at the patient level, with 80\% of patients used for training and 20\% for testing. Logistic models used feature standardization and class balancing. Since cognitive exploitation is a minority class, the classification threshold was tuned on the training set to maximize F1-score rather than fixed at 0.5:
\begin{equation}
\tau^{*}=\arg\max_{\tau\in\mathcal{T}} \mathrm{F1}(\tau).
\end{equation}
Performance was assessed using accuracy, precision, recall, F1-score, ROC-AUC, and average precision. The threshold-dependent measures were computed as
\begin{equation}
\mathrm{Precision}=\frac{TP}{TP+FP}, \quad
\mathrm{Recall}=\frac{TP}{TP+FN},
\end{equation}
\begin{equation}
\mathrm{F1}=\frac{2\cdot\mathrm{Precision}\cdot\mathrm{Recall}}{\mathrm{Precision}+\mathrm{Recall}}, \quad
\mathrm{Accuracy}=\frac{TP+TN}{TP+TN+FP+FN}.
\end{equation}
Because the target class is rare, precision-recall behavior and average precision were emphasized. A second evaluation restricted the test set to transactions inside vulnerability windows, which directly measures the framework’s intended safety benefit. Experiments were implemented in Python with \texttt{pandas}, \texttt{numpy}, and \texttt{scikit-learn}; the study should be interpreted as simulation-based validation rather than live deployment \cite{_8,_10}.

\section{Results and Discussion}

\subsection{Dataset Characteristics}

Table~\ref{tab:dataset_summary} summarizes the dataset. Cognitive exploitation accounted for 3.91\% of all transactions, and 9.76\% of transactions occurred during medication-induced vulnerability windows. The data are therefore imbalanced but still contain enough high-risk events for targeted vulnerability-window evaluation.

\begin{table}[htbp]
\centering
\caption{Summary statistics of the synchronized hybrid dataset}
\label{tab:dataset_summary}
\begin{tabular}{l c}
\hline
\textbf{Metric} & \textbf{Value} \\
\hline
Patients / days & 180 / 45 \\
Medication records / transactions & 8{,}100 / 30{,}855 \\
Missed dose rate & 0.1806 \\
Vulnerability-window transaction rate & 0.0976 \\
Cognitive exploitation rate & 0.0391 \\
Non-cognitive anomaly rate & 0.0341 \\
\hline
\end{tabular}
\end{table}

The dataset also reflects the intended clinical-financial coupling. Cognitive exploitation rose from 2.9\% outside vulnerability windows to 13.0\% inside vulnerability windows, while normal transactions still represented 82.5\% of vulnerability-window activity. Thus, the task is not solved by a simple rule that treats every missed-medication window as exploitative; models must still combine medication context with transaction-level features. Table~\ref{tab:vulnerability_distribution} reports this distribution.

\begin{table}[htbp]
\centering
\caption{Distribution of transaction labels by vulnerability status}
\label{tab:vulnerability_distribution}
\begin{tabular}{lccc}
\hline
\textbf{Vulnerability Status} & \textbf{Cognitive} & \textbf{Non-Cognitive} & \textbf{Normal} \\
\hline
Outside vulnerability window & 0.029 & 0.033 & 0.938 \\
Inside vulnerability window & 0.130 & 0.045 & 0.825 \\
\hline
\end{tabular}
\end{table}

\subsection{Feature-Level Findings}

The engineered features behaved consistently with the design assumptions. Vendor novelty was the strongest direct correlate of cognitive exploitation, followed by amount anomaly. Medication adherence showed a negative relationship with exploitation, indicating that missed medication was associated with higher risk but did not dominate the task. Frequency spike and time deviation had weaker marginal relationships, suggesting that they may be useful mainly in combination with medication state and other financial signals. This is important because a low marginal correlation does not imply that a feature is useless in an interaction model. A feature can contribute by changing the ranking of borderline cases during clinically sensitive periods, particularly when multiple weak cues appear together.

Class-wise feature means support this interpretation. Cognitive exploitation events had higher amount anomaly ($A=0.763$), universal vendor novelty ($V=1.000$), and lower medication adherence ($M=0.586$) than normal transactions ($A=0.244$, $V=0.113$, $M=0.824$). Non-cognitive anomalies occupied an intermediate profile, with elevated financial irregularity but less medication-related vulnerability. This separation indicates that the simulator preserves a meaningful difference between ordinary financial anomalies and cognitively risky exploitation events. Tables~\ref{tab:feature_summary} and~\ref{tab:feature_by_label} summarize these patterns.

\begin{table}[htbp]
\centering
\caption{Descriptive statistics of engineered features}
\label{tab:feature_summary}
\begin{tabular}{lccccc}
\hline
\textbf{Statistic} & \textbf{$A$} & \textbf{$V$} & \textbf{$F$} & \textbf{$T$} & \textbf{$M$} \\
\hline
Mean & 0.275 & 0.160 & 0.164 & 0.409 & 0.812 \\
Std & 0.347 & 0.367 & 0.312 & 0.365 & 0.391 \\
25\% & 0.000 & 0.000 & 0.000 & 0.100 & 1.000 \\
50\% & 0.000 & 0.000 & 0.000 & 0.100 & 1.000 \\
75\% & 0.569 & 0.000 & 0.000 & 0.600 & 1.000 \\
Max & 1.000 & 1.000 & 0.999 & 1.000 & 1.000 \\
\hline
\end{tabular}
\end{table}

\begin{table}[htbp]
\centering
\caption{Mean feature values by transaction label}
\label{tab:feature_by_label}
\begin{tabular}{lccccc}
\hline
\textbf{Label} & \textbf{$A$} & \textbf{$V$} & \textbf{$F$} & \textbf{$T$} & \textbf{$M$} \\
\hline
Cognitive exploitation & 0.763 & 1.000 & 0.224 & 0.479 & 0.586 \\
Financial anomaly non-cognitive & 0.555 & 0.481 & 0.202 & 0.520 & 0.745 \\
Normal & 0.244 & 0.113 & 0.160 & 0.402 & 0.824 \\
\hline
\end{tabular}
\end{table}

\begin{table}[htbp]
\centering
\caption{Correlation matrix of engineered features and target label}
\label{tab:correlation_matrix}
\begin{tabular}{lcccccc}
\hline
 & \textbf{$A$} & \textbf{$V$} & \textbf{$F$} & \textbf{$T$} & \textbf{$M$} & \textbf{Label} \\
\hline
$A$ & 1.000 & 0.413 & 0.115 & 0.186 & -0.240 & 0.283 \\
$V$ & 0.413 & 1.000 & 0.048 & 0.108 & -0.136 & 0.462 \\
$F$ & 0.115 & 0.048 & 1.000 & -0.134 & -0.093 & 0.039 \\
$T$ & 0.186 & 0.108 & -0.134 & 1.000 & 0.009 & 0.038 \\
$M$ & -0.240 & -0.136 & -0.093 & 0.009 & 1.000 & -0.117 \\
Label & 0.283 & 0.462 & 0.039 & 0.038 & -0.117 & 1.000 \\
\hline
\end{tabular}
\end{table}

Table~\ref{tab:correlation_matrix} confirms that vendor novelty and amount anomaly carry the strongest direct statistical association with exploitation, while medication adherence contributes as a weaker but clinically meaningful contextual signal. This pattern supports the paper’s design choice: medication is not expected to replace financial evidence, but to alter its interpretation when vulnerability is elevated.

\subsection{Global Model Comparison}

Table~\ref{tab:global_model_results} reports global performance on unseen patients. The financial-only baseline achieved the highest threshold-specific F1-score, while the additive medication model increased recall at the cost of precision. The weighted risk engine was interpretable but underperformed learned models, indicating that fixed heuristic weights are too rigid for the full task.

\begin{table*}[htbp]
\centering
\caption{Global performance comparison on unseen patients}
\label{tab:global_model_results}
\begin{tabular}{lcccccc}
\hline
\textbf{Model} & \textbf{Acc.} & \textbf{Prec.} & \textbf{Rec.} & \textbf{F1} & \textbf{AUC} & \textbf{AP} \\
\hline
Baseline Logistic $(A,V,F,T)$ & 0.9504 & 0.4407 & 0.5778 & 0.5000 & 0.9626 & 0.4279 \\
Additive Logistic $(A,V,F,T,M)$ & 0.9453 & 0.4098 & 0.6222 & 0.4941 & 0.9619 & 0.4258 \\
Weighted Risk Engine & 0.9456 & 0.3571 & 0.3333 & 0.3448 & 0.6521 & 0.1948 \\
Interaction Logistic & 0.9410 & 0.3870 & 0.6407 & 0.4826 & 0.9613 & 0.4476 \\
\hline
\end{tabular}
\end{table*}

The interaction-aware model did not exceed the baseline in global F1-score, but it achieved the highest recall and average precision. It recovered 173 true positives compared with 156 for the financial-only baseline, while false positives increased from 198 to 274. This trade-off is expected in a safety-oriented detector: the interaction model is more aggressive and better suited for prioritizing risky cases, but it should be paired with graduated interventions such as caregiver review or secondary authentication rather than automatic blocking. Tables~\ref{tab:global_confusion} and~\ref{tab:ablation} make the trade-off explicit.

\begin{table}[htbp]
\centering
\caption{Confusion matrix breakdown for global evaluation}
\label{tab:global_confusion}
\begin{tabular}{lcccc}
\hline
\textbf{Model} & \textbf{TN} & \textbf{FP} & \textbf{FN} & \textbf{TP} \\
\hline
Baseline Logistic $(A,V,F,T)$ & 5818 & 198 & 114 & 156 \\
Additive Logistic $(A,V,F,T,M)$ & 5774 & 242 & 102 & 168 \\
Interaction Logistic & 5742 & 274 & 97 & 173 \\
Weighted Risk Engine & 5854 & 162 & 180 & 90 \\
\hline
\end{tabular}
\end{table}

\begin{table}[htbp]
\centering
\caption{Ablation study of medication-aware components}
\label{tab:ablation}
\begin{tabular}{lcccc}
\hline
\textbf{Model} & \textbf{F1} & \textbf{Recall} & \textbf{Precision} & \textbf{AP} \\
\hline
Financial Only $(A,V,F,T)$ & 0.5000 & 0.5778 & 0.4407 & 0.4279 \\
$+M$ Additive Medication & 0.4941 & 0.6222 & 0.4098 & 0.4258 \\
$+M$ Interaction Terms & 0.4826 & 0.6407 & 0.3870 & 0.4476 \\
\hline
\end{tabular}
\end{table}

\subsection{Interaction-Aware Interpretation}

The coefficient pattern supports the claim that medication works best as a contextual modifier. Vendor novelty remained the dominant signal, with coefficient 3.6561, while amount anomaly had coefficient 0.5103. Among interaction terms, $V(1-M)$ was positive at 0.3974 and $A(1-M)$ was positive at 0.0953. Therefore, unfamiliar vendors and abnormal amounts became more suspicious when medication was missed. The direct medication coefficient was also positive in the standardized model, but the interaction terms provide the central interpretation: clinical context changes how financial anomalies should be read. Table~\ref{tab:interaction_coefficients} lists the learned coefficients.

\begin{table}[htbp]
\centering
\caption{Coefficients of the interaction-aware logistic model}
\label{tab:interaction_coefficients}
\begin{tabular}{lc}
\hline
\textbf{Feature} & \textbf{Coefficient} \\
\hline
$V$ & 3.6561 \\
$A$ & 0.5103 \\
$M$ & 0.4420 \\
$V(1-M)$ & 0.3974 \\
$T$ & 0.1177 \\
$A(1-M)$ & 0.0953 \\
$F$ & 0.0951 \\
$F(1-M)$ & -0.0885 \\
$T(1-M)$ & -0.2427 \\
\hline
\end{tabular}
\end{table}

This result clarifies the difference between additive and interaction-aware modeling. Adding medication directly increases recall, but interaction modeling better aligns with the clinical hypothesis. Missed medication alone should not be treated as exploitation; instead, it should raise the priority of financial irregularities that occur during a cognitively vulnerable period.

\subsection{Precision-Recall and Vulnerability-Window Evaluation}

\begin{figure}[htbp]
    \centering
    \includegraphics[width=0.82\linewidth]{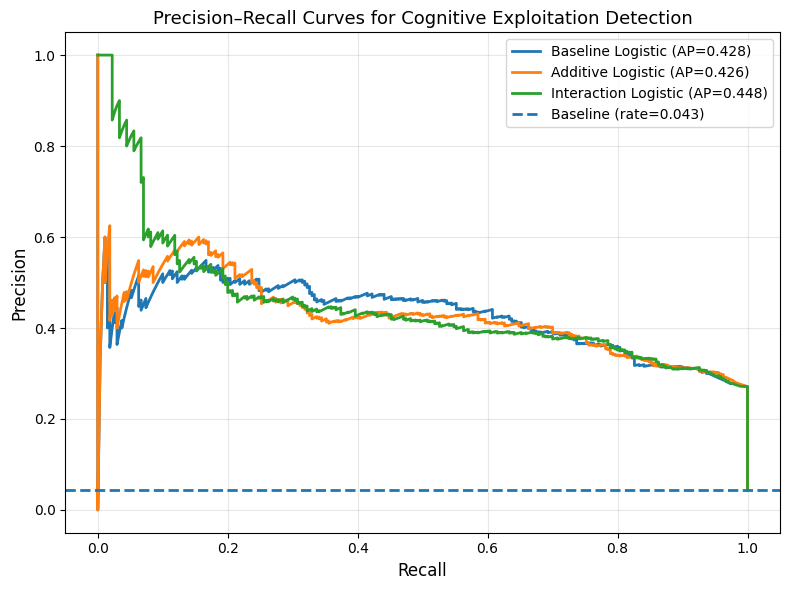}
    \caption{Precision-recall curves for financial-only, additive medication-aware, and interaction-aware models. The interaction-aware model achieves the highest average precision, indicating improved ranking quality for high-risk exploitation cases.}
    \label{fig:pr_curve}
\end{figure}

Because cognitive exploitation is rare, precision-recall analysis is more informative than accuracy alone. Figure~\ref{fig:pr_curve} shows that the interaction-aware model achieved the highest average precision, indicating better ranking of high-risk cases across thresholds.

The main safety-relevant result appears in vulnerability-window evaluation. Table~\ref{tab:vulnerability_eval} compares the financial-only baseline and interaction-aware model using only transactions that occurred during missed-medication vulnerability windows.

\begin{table}[htbp]
\centering
\caption{Performance during vulnerability windows only}
\label{tab:vulnerability_eval}
\begin{tabular}{lccc}
\hline
\textbf{Model} & \textbf{Precision} & \textbf{Recall} & \textbf{F1} \\
\hline
Baseline Logistic $(A,V,F,T)$ & 0.4923 & 0.7442 & 0.5926 \\
Interaction Logistic & 0.4333 & 0.9070 & 0.5865 \\
\hline
\end{tabular}
\end{table}

The interaction-aware model increased recall from 0.7442 to 0.9070 during vulnerability windows. Precision decreased from 0.4923 to 0.4333, and F1 dropped only slightly from 0.5926 to 0.5865. For the target application, this recall gain is important because missed exploitation events during cognitively vulnerable periods are more harmful than additional cases sent for review. The result shows that the proposed model should not be viewed as a universal replacement for financial anomaly detection, but as a vulnerability-sensitive safeguard that becomes especially useful when clinical context indicates elevated risk.

The global and vulnerability-window results should therefore be interpreted together. Globally, the financial-only baseline remains strong because vendor novelty and amount anomaly already capture much of the exploitation signal. Inside vulnerability windows, however, the interaction model better reflects the clinical objective of minimizing missed harmful events. This explains why the proposed model may appear slightly weaker under a single global F1 operating point while still being preferable for safety-critical triage.

\subsection{Practical Implications and Limitations}

The findings support a layered protection strategy rather than a single automatic decision rule. A financial-only detector can operate as a general screening mechanism, while the interaction-aware model can increase sensitivity during medication-linked vulnerability windows. In practice, low-risk transactions could proceed normally, moderate-risk cases could trigger step-up verification, and high-risk transactions during missed-medication periods could be routed to a trusted caregiver, financial guardian, or bank officer for review. This graduated approach is important because older adults with cognitive impairment may still retain meaningful financial autonomy, and unnecessary blocking could reduce independence or create frustration. The goal is therefore proportionate intervention, consistent with human-centered protection principles \cite{_15}.

The framework also highlights the importance of privacy and explainability. Medication adherence and transaction records are sensitive, so a deployable system should avoid centralizing raw medication histories, detailed banking logs, or identity-rich behavioral traces whenever possible. A local device or trusted gateway could instead compute compact indicators such as missed-dose status, vendor novelty, burst score, abnormal amount, and risk tier. Only the alert and a minimal explanation would need to be escalated. Such explanations should clarify why a transaction was flagged, for example, an unfamiliar vendor and abnormal amount occurring during a missed-medication vulnerability window, without presenting the alert as a clinical diagnosis.

Several limitations remain. The dataset is simulated, so the results should be interpreted as controlled validation of the modeling concept rather than evidence of immediate clinical or banking deployment readiness. The weighted risk engine used fixed parameters, which improves interpretability but limits adaptability. The interaction-aware model also increased false positives, meaning that real-world use would require careful threshold selection, consent procedures, auditability, calibration across patient groups, and human-in-the-loop review. The simulation design nevertheless preserves an important distinction between financial anomalies and cognitive exploitation by allowing normal transactions inside vulnerability windows and non-cognitive anomalies outside them.

Future work should test the framework using real medication adherence and transaction traces, patient-specific adaptation, adaptive thresholds, privacy-preserving edge implementations, and caregiver-centered escalation workflows. The proposed simulation structure supports reproducibility because individual components, such as medication simulation, vendor generation, or threshold selection, can be replaced without changing the full evaluation framework. Future studies should also assess whether medication-aware explanations improve caregiver, clinician, or bank analyst review quality, including response time, trust, perceived burden, and willingness to act. These steps are necessary to translate medication-aware financial protection from simulation into responsible practice.

\section{Conclusion}

This paper presented a medication-aware financial exploitation detection framework for Alzheimer’s patients by integrating medication adherence with transaction-level financial monitoring. The results show that financial-only models remain competitive globally, but medication-aware interaction modeling is more effective during clinically elevated vulnerability windows. In particular, the interaction-aware model increased vulnerability-window recall from 0.7442 to 0.9070, indicating substantially improved sensitivity to cognitively risky transactions.

The central finding is that medication adherence is most useful not as an isolated predictor, but as a contextual signal that changes how financial anomalies should be interpreted. Although the study is limited to simulation-based validation, it provides evidence that synchronizing clinical and financial signals can support more context-aware protection for cognitively vulnerable individuals.

\bibliographystyle{splncs04}
\bibliography{references}

@article{_1,
  author    = {Ebner, Natalie C. and Pehlivanoglu, Didem and Shoenfelt, Alayna},
  title     = {Financial Fraud and Deception in Aging},
  journal   = {Advances in Geriatric Medicine and Research},
  year      = {2023},
  volume    = {5},
  number    = {3},
  pages     = {e230007},
  doi       = {10.20900/agmr20230007}
}

@article{_2,
  author    = {Mazzonna, Fabrizio and Peracchi, Franco},
  title     = {Are Older People Aware of Their Cognitive Decline? Misperception and Financial Decision-Making},
  journal   = {Journal of Political Economy},
  year      = {2024},
  volume    = {132},
  number    = {6},
  pages     = {1793--1830},
  doi       = {10.1086/728697}
}

@article{_3,
  author    = {Yang, Ziqi and Xu, Xuhai and Yao, Bingsheng and Rogers, Ethan and Zhang, Shao and Intille, Stephen S. and Shara, Nawar and Gao, Guodong Gordon and Wang, Dakuo},
  title     = {Talk2Care: An {LLM}-based Voice Assistant for Communication between Healthcare Providers and Older Adults},
  journal   = {Proceedings of the ACM on Interactive, Mobile, Wearable and Ubiquitous Technologies},
  year      = {2024},
  volume    = {8},
  number    = {2},
  pages     = {73:1--73:35},
  doi       = {10.1145/3659625}
}

@inproceedings{_4,
  author    = {Zhang, Kai and Kang, Yangyang and Zhao, Fubang and Liu, Xiaozhong},
  title     = {{LLM}-based Medical Assistant Personalization with Short- and Long-Term Memory Coordination},
  booktitle = {Proceedings of the 2024 Conference of the North American Chapter of the Association for Computational Linguistics: Human Language Technologies (Volume 1: Long Papers)},
  year      = {2024},
  pages     = {2386--2398},
  address   = {Mexico City, Mexico},
  publisher = {Association for Computational Linguistics},
  doi       = {10.18653/v1/2024.naacl-long.132}
}

@article{_5,
  author    = {Ud Din, Fareed and Giri, Nabaraj and Shetty, Namrata and Hilton, Tom and Shafiabady, Niusha and Tully, Phillip J.},
  title     = {Co-Designing a {DSM-5}-Based {AI}-Powered Smart Assistant for Monitoring Dementia and Ongoing Neurocognitive Decline: Development Study},
  journal   = {BioMedInformatics},
  year      = {2025},
  volume    = {5},
  number    = {3},
  pages     = {49},
  doi       = {10.3390/biomedinformatics5030049}
}

@article{_6,
  author    = {Shaik, Mohmmad Arif and Anik, Fahim Islam and Hasan, Md. Mehedi and Chakravarty, Sumit and Ramos, Mary Dioise and Rahman, Mohammad Ashiqur and Ahamed, Sheikh Iqbal and Sakib, Nazmus},
  title     = {Advancing Remote Monitoring for Patients With Alzheimer Disease and Related Dementias: Systematic Review},
  journal   = {JMIR Aging},
  year      = {2025},
  volume    = {8},
  pages     = {e69175},
  doi       = {10.2196/69175}
}

@article{_7,
  author    = {Anik, Fahim Islam and Hasan, Md Mehedi and Rodriguez-Cardenas, Juan and Beiswanger, Richard and Tasnim, Masrura and Ramos, Mary and Sakib, Nazmus},
  title     = {{IoMT} and Data Privacy in Alzheimer{\textquoteright}s Care for Older Adults: A Systematic Review},
  journal   = {EAI Endorsed Transactions on Pervasive Health and Technology},
  year      = {2025},
  volume    = {11},
  number    = {1},
  doi       = {10.4108/eetpht.11.6170}
}

@article{_8,
  author    = {Dall'Ora, Nicola and Felli, Lorenzo and Aldegheri, Stefano and Vicino, Nicola and Giuliano, Romeo},
  title     = {LumiCare: A Context-Aware Mobile System for Alzheimer{\textquoteright}s Patients Integrating {AI} Agents and {6G}},
  journal   = {Electronics},
  year      = {2025},
  volume    = {14},
  number    = {17},
  pages     = {3516},
  doi       = {10.3390/electronics14173516}
}

@misc{_9,
  author       = {Bazgir, Adib and Habibdoust, Amir and Song, Xing and Zhang, Yuwen},
  title        = {AgenticAD: A Specialized Multiagent System Framework for Holistic Alzheimer Disease Management},
  year         = {2025},
  eprint       = {2510.08578},
  archivePrefix= {arXiv},
  primaryClass = {cs.AI},
  url          = {https://arxiv.org/abs/2510.08578}
}

@article{_10,
  author    = {Grammenos, Gerasimos and Vrahatis, Aristidis G. and Lazaros, Konstantinos and Exarchos, Themis P. and Vlamos, Panagiotis and Krokidis, Marios G.},
  title     = {AI Agents in Alzheimer{\textquoteright}s Disease Management: Challenges and Future Directions},
  journal   = {Frontiers in Aging Neuroscience},
  year      = {2026},
  volume    = {17},
  pages     = {1735892},
  doi       = {10.3389/fnagi.2025.1735892}
}

@article{_11,
  author    = {Olchanski, N. and Daly, A. T. and Zhu, Y. and Breslau, R. and Cohen, J. T. and Neumann, P. J. and others},
  title     = {Alzheimer's Disease Medication Use and Adherence Patterns by Race and Ethnicity},
  journal   = {Alzheimer's \& Dementia},
  year      = {2023},
  volume    = {19},
  number    = {4},
  pages     = {1184--1193},
  doi       = {10.1002/alz.12753}
}

@article{_12,
  author    = {Mishra, Pratik K. and Iaboni, Andrea and Ye, Bing and Newman, Kristine and Mihailidis, Alex and Khan, Shehroz S.},
  title     = {Privacy-Protecting Behaviours of Risk Detection in People with Dementia Using Videos},
  journal   = {BioMedical Engineering OnLine},
  year      = {2023},
  volume    = {22},
  number    = {1},
  pages     = {4},
  doi       = {10.1186/s12938-023-01065-3}
}

@article{_13,
  author    = {Giannouli, Vaitsa},
  title     = {Can Changes in Financial Performance Be Used in the Diagnosis of Neurocognitive Disorders? A Systematic Review of Findings from Greece},
  journal   = {Brain Sciences},
  year      = {2024},
  volume    = {14},
  number    = {11},
  pages     = {1113},
  doi       = {10.3390/brainsci14111113}
}

@article{_14,
  author    = {Trendl, Anna and Anwyl-Irvine, Alexander and Vomfell, Lara and Abbey, Emma and Stewart, Neil and Atkins, David and Llewellyn, David J. and Gathergood, John and Leake, David},
  title     = {Early Behavioral Markers of Loss of Financial Capacity},
  journal   = {JAMA Network Open},
  year      = {2025},
  volume    = {8},
  number    = {6},
  pages     = {e2515894},
  doi       = {10.1001/jamanetworkopen.2025.15894}
}

@article{_15,
  author    = {Zhai, Yuxiang and Xue, Xiao and Guo, Zekai and Jin, Tongtong and Diao, Yuting and Jeung, Jihong},
  title     = {Hear Us, then Protect Us: Navigating Deepfake Scams and Safeguard Interventions with Older Adults through Participatory Design},
  journal   = {Proceedings of the 2025 CHI Conference on Human Factors in Computing Systems},
  year      = {2025},
  pages     = {1--19},
  doi       = {10.1145/3706598.3714423}
}

@article{_16,
  author    = {Wei, Wenxing and Balser, Sarah},
  title     = {A Scoping Review: Financial Exploitation Among Older People Living with Dementia},
  journal   = {Trauma, Violence, \& Abuse},
  year      = {2025},
  doi       = {10.1177/15248380251383930}
}

@inproceedings{al2025bayesian,
  title={Bayesian modeling for uncertainty management in financial risk forecasting and compliance},
  author={Al Mamun, Sharif and Hossain, Rakib and Rahman, Md Jobayer and Devnath, Malay Kumar and Afroz, Farhana and Al Amin, Lisan},
  booktitle={2025 IEEE International Conference on Data Mining Workshops (ICDMW)},
  pages={178--186},
  year={2025},
  organization={IEEE}
}

@inproceedings{hossain2025risk,
  title={Risk-Controlled Multimodal Emotion Coaching for Autism Support Using Self-Supervised Vision and Speech Encoders},
  author={Hossain, Rakib and Ali, Lasker Ershad and Ripon, Kazi Shah Nawaz},
  booktitle={2025 40th International Conference on Image and Vision Computing New Zealand (IVCNZ)},
  pages={1--7},
  year={2025},
  organization={IEEE}
}

@inproceedings{akter2026hybrid,
  title={Hybrid Federated and Split Learning for Privacy Preserving Clinical Prediction and Treatment Optimization},
  author={Akter, Farzana and Hossain, Rakib and Toushi, Deb Kanna Roy and Khan, Mahmood Menon and Amin, Sultana and Al Amin, Lisan},
  booktitle={SoutheastCon 2026},
  pages={1--8},
  year={2026},
  organization={IEEE}
}

@inproceedings{hossain2025profit,
  title={Profit-Optimal Machine Learning for Purchase Intent Prediction: Calibration, Temporal Validation, and Economic Constraints},
  author={Hossain, Rakib and Ali, Lasker Ershad and Ripon, Kazi Shah Nawaz and Bulbul, Md Farhad and Tonmoy, Md Mehedi Hasan},
  booktitle={2025 28th International Conference on Computer and Information Technology (ICCIT)},
  pages={3087--3092},
  year={2025},
  organization={IEEE}
}

@article{al2022data,
  title={Data driven classification of opioid patients using machine learning--an investigation},
  author={Al Amin, Lisan and Mukta, Md Saddam Hossain and Saikat, Md Sezan Mahmud and Hossain, Md Ismail and Islam, Md Adnanul and Ahmed, Mohiuddin and Azam, Sami},
  journal={IEEE Access},
  volume={11},
  pages={396--409},
  year={2022},
  publisher={IEEE}
}

\end{document}